# A Vision System for Multi-View Face Recognition

M. Y. Shams  
Mansoura University,  
Computer Science Department  
myysd2016@mans.edu.eg

A. S. Tolba  
Mansoura University,  
Computer Science Department  
astolba@mans.edu.eg

S. H. Sarhan  
Mansoura University,  
Computer Science Department  
sh_sarhan@mans.edu.eg

*Abstract-* Multimodal biometric identification has been grown a great attention in the most interests in the security fields. In the real world there exist modern system devices that are able to detect, recognize, and classify the human identities with reliable and fast recognition rates. Unfortunately most of these systems rely on one modality, and the reliability for two or more modalities are further decreased. The variations of face images with respect to different poses are considered as one of the important challenges in face recognition systems. In this paper, we propose a multimodal biometric system that able to detect the human face images that are not only one view face image, but also multi-view face images. Each subject entered to the system adjusted their face at front of the three cameras, and then the features of the face images are extracted based on Speeded Up Robust Features (SURF) algorithm. We utilize Multi-Layer Perceptron (MLP) and combined classifiers based on both Learning Vector Quantization (LVQ), and Radial Basis Function (RBF) for classification purposes. The proposed system has been tested using SDUMLA-HMT, and CASIA datasets. Furthermore, we collected a database of multi-view face images by which we take the additive white Gaussian noise into considerations. The results indicated the reliability, robustness of the proposed system with different poses and variations including noise images.

## I. Introduction

THE most recent researchers in biometrics are trying to overcome the challenges of such systems including the implementation of large scale (*Big Data*) biometrics, the problems of invariant face recognition, dealing with sparse data representation, and when the biometrics merit the internet of things.

In this paper, we present a real time system that able to detect, recognize, and classify face images using three cams as shown in Figure (1). The system starting with capturing the face image (acquisition stage), then the preprocessing stage is applied to get rid of background and to transform RGB to gray scale level. The laptop camera may suffer from resolution, producing a low quality image so we are seeking to overcome the pixel resolution problem using spatial domain filtering (mean filter). The features are extracted using SURF algorithm including the analysis of face image. The architecture of the proposed system relies on parallel fusion of the acquired face images with different poses and variation angles. The evaluation results are based on using a standard validated database of SDUMLA-HMT and CASIA datasets. Furthermore, with the aid of our colleague and students, we have collected a database consists of 50 male and 20 female that are enrolled to the system simultaneously with validated mentioned standard databases. We tested the proposed algorithm using different angles (0º, 45º, and 90º) and the results shows that at front camera with 0º the Genuine Acceptance Rate (GAR) are increased and it is slightly decreased to reaches to 90º. The main contributions of this paper are listed as follows:

- We implement a real time multimodal biometric system based on simultaneous acquisition of face images with different poses using three cams.
- We utilizes SURF algorithm to extract the invariant face images with different poses and views degrees.
- The proposed system has been tested based on Multi-Layer Perceptron (MLP) and combined classifiers for evaluating the GAR of the enrolled face templates.
- We add additive white Gaussian noise in order to evaluate the performance of the system in the presence of low quality image and noise.

The organization of the paper conducted the related work in section II. The methodologies of the proposed system are illustrated in section III. The evaluation results are investigated in section IV. Finally the conclusion and future work in section V.

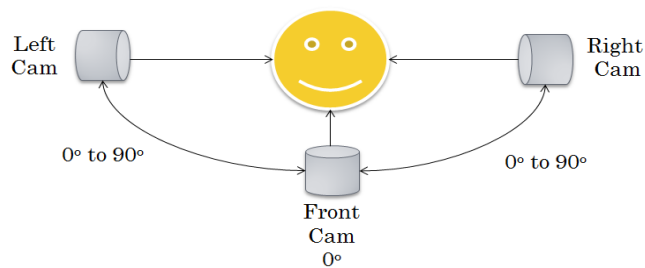

Fig. 1: Acquiring face image using three cams.

## II. Related work

A survey of the approaches and challenges for face recognition system are illustrated in [1]. A. Wanger et al [2] proposed a practical face recognition system based on multi-sample face images with different alignment and illumination using sparse representation. D. Chen et al [3] presented an algorithm used the high dimensional features and compression based on Local Binary Pattern (LBP) and a rotated sparse regression of face images. In some cases we need to recognize a partial of face images as introduced by S. Liao et al [4] by which they used a patch face images based on Scale Invariant Feature Transform (SIFT) algorithm and Gabor Ternary Pattern (GTP) descriptor. Joint sparse dynamic features for multi-sample view face images are presented by H. Zhang et al [5]. A Gabor features are extracted for representing the face images in the presence of occlusion sparse representation is introduced by M. Yang [6]. Y. Xu et al [7] proposed an approach to exploit the limited training samples of two–step face recognition using FERET datasets.

A system based on multi-attribute spaces for face images depending on calibration and similarity search algorithm is introduced by W. Scheirer et al [8]. With the varying of poses, expressions, and illumination using Kinect query face image presented by B. Li et al [9]. They used Sparse Representation Classifier (SRC) for face recognition by formulating the sparse coding of the query images. O. Barkan et al [10] presented a fast high dimensional vector multiplication for face recognition using Over-Completes Local Binary Pattern (OCLBP). They used Linear Discriminant Analysis (LDA) and class covariance normalization for feature extraction and matching. A multi-view face detection technique based on Deep Convolutional Neural Networks (DCNN) with different angles and rotation is presented by [11]. C. Ding et al [12] proposed an algorithm for face recognition system using DCNN and three-layer Stacked Auto-Encoder (SAE), and they achieves a higher recognition rate with different poses and occlusion scenarios. A short paper presented by M. Kan et al [13] discussed the multi-view discriminant analysis applied on face images taking the tasks of face recognition including pose estimation. A face recognition based on multi-directional multi-level dual-cross patterns is presented by C. Ding et al [14]. Automatic face normalization approach for single sample face recognition is presented by M. Haghighat et al [15]. They used the Histogram of Oriented Gradients (HOG) for feature extraction and matching using Canonical Correlation Analysis (CCA). A pose-aware face recognition based on CNN to learn the pose variation is proposed by I. Masi et al [16]. In this paper, we are seeking to recognize each subject enrolled to the system with different poses and angles. Depending on the collected database and the standard validated databases, we evaluate the performance of the proposed system in presence of additive white Gaussian noise.

III. PROPOSED ALGORITHM

Researches in biometrics and computer vision are tried to enhance the face recognition systems by considering face variables including pose variation, illumination, aging, noisy face images, and occlusion problems. In this paper, we present a robust real time multi-view face recognition system based on SURF algorithm, and neural network approaches. We utilized a well-known neural networks; Multi-Layer Perceptron (MLP), Combined LVQ, and RBF classifiers in order to evaluate the results of the proposed system. The decision of the system is based on rank, and decision level fusion for each output classifiers as shown in Figure (2).

Each subject enrolled to the system acquires multi-view sample face images based on SDUMLA-HMT [17], CASIA [18], and our collected face database. We used three cameras adjusted as shown in Figure (1) with different poses and angles. Figure (3) shows the samples of face images with different poses and variations of the mentioned databases.

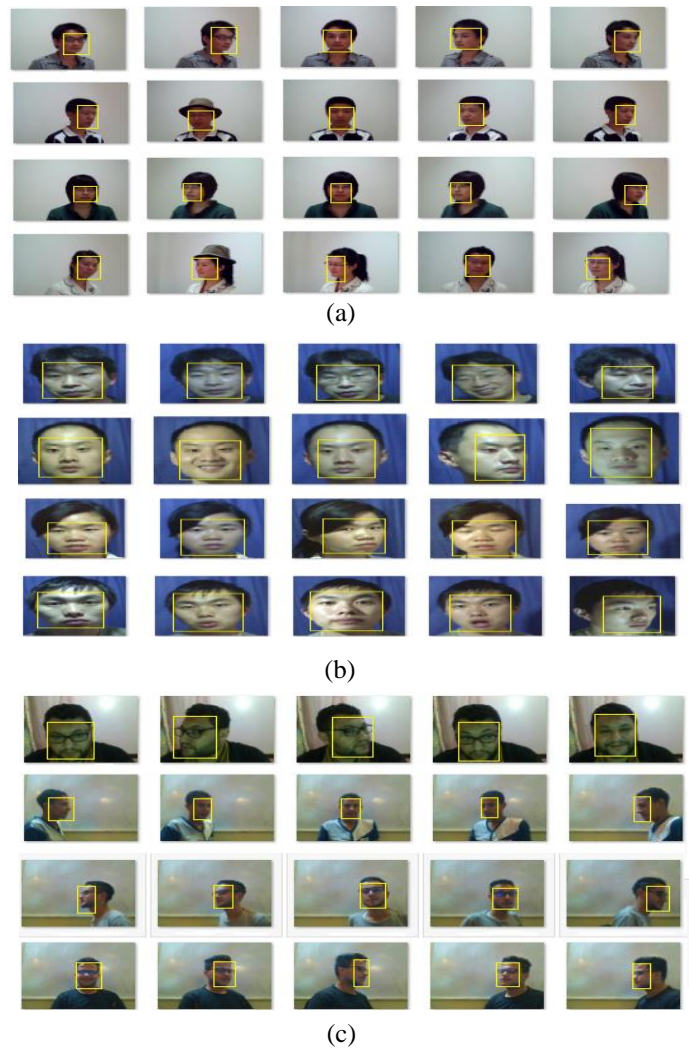

Fig 3: Multi-view face samples of (a) SDUMLA-HMT, (b) CASIA-V5, and (c) Our collected datasets.

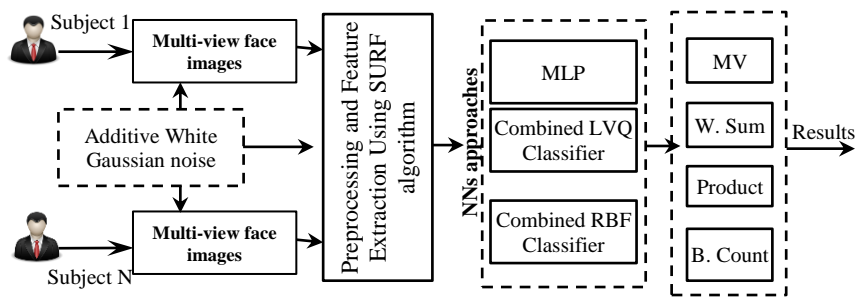

Fig 2: General Framework of the proposed system.

*A. SURF Algorithm*

Speeded Up Robust Features (SURF) algorithm is considered as a one of the robust and effective algorithms used for face detection and recognition. SURF algorithm is faster than Scale Invariant Feature Transform (SIFT) by three times with a high recall precision not worse than SIFT. That is more effective for handling the blurring and rotation problems [19].

Detection based on SURF algorithm depends up on Hessian interest point localization given by H as follows:

$$H = \begin{vmatrix} L_{xx} & L_{xy} \\ L_{xy} & L_{yy} \end{vmatrix} \quad (1)$$

Where $L_{xx}$ is the Laplacian Gaussian of the face image. H is determined by the convolution of the Gaussian second order derivative with the enrolled face images. In order to approximate $L_{xx}$, $L_{xy}$, and $L_{yy}$, we use $D_{xx}$, $D_{yy}$, and $D_{xy}$ which is determined by

$$det(H_{approx}) = D_{xx} D_{yy} - (wD_{xy})^2 \quad (2)$$

Where $w$ are the weights of the rectangular region.

In this paper, we propose an algorithm that able to overcome the problems of additive noises generated by different sources caused by: Camera sensor, detector sensitivity variation, surrounding environmental effects, discrete nature of radiation, dust on the optics, quantization errors, and/or transmission data errors [20]. The noise can be considered as an additive white Gaussian noise $N(i,j)$ with zero mean and variance $\sigma$, therefore the enrolled image $I(i,j)$ is the sum of the true image $T(i,j)$ and the noise-image $N(i,j)$ as in equation (3)

$$I(i,j) = T(i,j) + N(i,j) \quad (3)$$

For multi-view face images, we determine the sum of the gray scale pixel values of the enrolled face images and the intermediate representation that are the integral image for major speeded up as follows:

$$I_{\sum(x)} = \sum_{i=0}^{i \leq x} \sum_{j=0}^{j \leq y} I(i,j) \quad (4)$$

Matching the fast indexing through the sign of the Laplacian for underlying interested parts is given by the trace T of the Hessian matrix in [19]:

$$T = L_{xx} + L_{yy} \quad (5)$$

G. Du et al [21] proposed an algorithm for using SURF algorithm in order to detect face images using 64 and 128 SIFT features and the results indicated that for 128-dimensional SIFT at the matching step, the SURF algorithm consumes less time than 64-dimensional features. In this paper we extract the face features using 128-dimensional features and the features are applied to the classification stage using MLP, combined LVQ, and RBF Classifier.

*B. Multi-Layer Perceptron*

We conducted a robust boosted parameter for rotation invariant texture in classification based on MLP and combined classifier [22]. A constructive training algorithm for MLP is presented by [23]. They developed an algorithm that used a single hidden layer for a given number of neurons and a small number of training patterns applied to facial expression recognition system. In this paper, we uses MLP ensemble classifier based on N input patterns extracted features from SURF algorithm. Therefore, we calculate the diversity matrix of each produced input pattern. The decision is based on majority voting (MV), weighted sum (W. Sum), product, and Borda count (B. Count).

*C. Combined LVQ, and RBF Classifier*

The classification of the extracted features generated by SURF algorithm is applied using combined LVQ, and RBF classifier. The architecture of learning vector quantization (LVQ) based on the combined classifier parameter is introduced by A. S. Tolba [24]. Furthermore, A. S. Tolba and A. N. Abu-Rezq [25] used both the combined LVQ (CLVQ) and combined radial basis function (CRBF) classifiers for measuring the performance of invariant face images. In this paper we obey the previous mentioned algorithms to classify the extracted features from SURF to ensure the superiority of our proposed system to overcome the poses, illumination, and noise problems.

IV. EVALUATION RESULTS

In this paper, we use three datasets that are SDUMLA-HMT, CASIA, and our collected database. SDUMLA-HMT consists of 8904 face images, we collect 424 face images with different poses and variations, and we took 318 face images for training and the remaining 106 for testing phase. CASIA consists of 2500 face images, we took 1500 face images for training and the remaining 1000 for testing. Our collected datasets consists of 70 subjects 50 male and 20 female. We collected 350 images with different poses and angle variation. We took 210 images for training and 140 images for testing. In [26] we present a combined LVQ classifier algorithm with variable parameters. In this paper we use the network parameters with fixed values during the experimental evaluation results as shown in Table 1. The network parameters include the number of hidden neurons, number of epoch's iterations, and the learning rate. The evaluation results are performed based on three cases that are listed as follows:

- **Case 1:** The Genuine acceptance rate (GAR) of the enrolled templates at front of camera with $0^o$ (see Table 2).
- **Case 2:** The GAR of the enrolled templates with different poses and angles variations (Table 3).
- **Case 3:** The GAR after adding the additive white Gaussian noise with zero mean and different variances (σ) (Figure. 4). For case 3: The noise can be handled using whether spatial domain filtering or frequency domain filtering.

In this paper, we used spatial domain filter that refer to the enrolled face image based on direct manipulation of image pixels. We used linear spatial domain filter also called mean filter for noise removal from the face image. The performance evaluation depends on computing a single number that reflects the recognition rate of the probe face images out of the gallery images taking the following metrics into consideration [20]:

### A. Mean Square Error

The mean square error (MSE) is the difference between the probe face image and the gallery images. Consider $I(i,j)$ is the additive noise, $T(i,j)$ is the true image, and $N(i,j)$ is the noise image as given in equation (3). Then the MSE is determined as in equation (6).

$$MSE = \frac{1}{N^2}\sum_{i=1}^{N}\sum_{j=1}^{N}[I(i,j) - G(i,j)]^2 \quad (6)$$

Where N is the image size, $I(i,j)$ is the probe image, and $G(i,j)$ is the gallery image. Note that the probe image including the additive white Gaussian noise. The MSE will decrease as long as noise reaches to zero and the difference between probe and gallery image reaches to zero.

### B. Root Mean Square Error

The root mean square error (RMSE) will increase when the probe face image deviate from the gallery image as in equation (7).

$$RMSE = \sqrt{\frac{1}{N^2}\sum_{i=1}^{N}\sum_{j=1}^{N}[I(i,j) - G(i,j)]^2} \quad (7)$$

### C. Mean Absolute Error

The mean absolute error (MAE) is determined as in equation (8):

$$MSE = \frac{1}{N^2}\sum_{i=1}^{N}\sum_{j=1}^{N}|I(i,j) - G(i,j)| \quad (8)$$

### D. Percentage Fit Error

The percentage fit error (PFE) is determined in equation (9)

$$PFE = \frac{norm(I(i,j) - G(i,j))}{norm(I(i,j))} \times 100 \quad (9)$$

### E. Signal –to – Noise Ratio

The signal-to-noise factor is considered as one of the important factors for measuring the quality of the probe image in presence of noise. The more SNR value, the more quality and free noise are obtained. The SNR is computed in equation (10):

$$SNR = 10\log_{10}\left\lfloor \frac{\sum_{i=1}^{N}\sum_{j=1}^{N}[I(i,j)]^2}{\sum_{i=1}^{N}\sum_{j=1}^{N}[I(i,j) - G(i,j)]^2} \right\rfloor \quad (10)$$

### F. Peak Signal –to – Noise Ratio

The peak signal-to-noise (PSNR) is computed as in equation (11) given the maximum pixel intensity $I_{max}$.

$$SNR = 10\log_{10}\left\lfloor \frac{I_{max}^2}{\sum_{i=1}^{N}\sum_{j=1}^{N}[I(i,j) - G(i,j)]^2} \right\rfloor \quad (11)$$

TABLE 1: THE NEURAL NETWORK PARAMETERS FOR MLP, COMBINED LVQ, AND RBF CLASSIFIER.

| Parameter | Values |
| --- | --- |
| Number of hidden neurons | 500 |
| Number of epochs (iterations) | 300 |
| Learning rate | 0.03 |

TABLE 2: THE GAR (%) AT THE FRONTAL CAMERA WITH ZERO DEGREE.

| Data Base | SURF + (MLP, CLVQ, and CRBF) | GAR (%) | | | |
| --- | --- | --- | --- | --- | --- |
| | | MV | W.Sum | Prod | B. Count |
| SDUMLA-HMT | MLP | 93.35 | 92.24 | 91.54 | 90.25 |
| | CLVQ | 96.54 | 95.24 | 94.57 | 94.00 |
| | CRBF | 92.25 | 90.64 | 89.34 | 87.01 |
| CASIA | MLP | 92.01 | 90.25 | 91.65 | 90.31 |
| | CLVQ | 95.65 | 94.45 | 93.64 | 93.50 |
| | CRBF | 94.78 | 93.25 | 92.01 | 91.03 |
| Our Collected Database | MLP | 94.56 | 93.80 | 92.09 | 92.07 |
| | CLVQ | 97.25 | 96.34 | 95.65 | 94.55 |
| | CRBF | 94.99 | 94.25 | 94.35 | 93.15 |

We found that the majority voting (MV) algorithm achieves higher recognition rates than weighted sum (W. Sum), Product (Prod), and Borda count (B. count). As shown in Figure (1) by changing the position of the camera by the range from $0^o$ to $90^o$, we obtain different face views. Noted that the frontal camera is moved and both right and left camera are fixed position. Table 3 shows the results obtained with the angle variation $45^o$ to the right cam and $45^o$ to the left cam. By adding additive white Gaussian noise to evaluate our proposed system for handling the noise effect, Figure (4) shows the evaluation metrics of MSE, RMSE, MAE, PFE, SNR (dB), and PSNR (dB) with different noise variances (σ) with interval [0 0.1]. The experimental results performances based on adding Gaussian noise are based on our collected database. The GAR of the proposed system in presence of additive white Gaussian noise is shown in Table 4 for our collected database.

TABLE 3: THE GAR (%) OF THE PROPOSED SYSTEM WITH MULTI-VIEWS ANGLES.

| Data Base | SURF + (MLP, CLVQ, and CRBF) | GAR (%) of View 1 for 45º Left Pose | | | | GAR (%) of View 2 for 45º Right Pose | | | |
|---|---|---|---|---|---|---|---|---|---|
| | | MV | W.Sum | Prod | B. Count | MV | W.Sum | Prod | B. Count |
| SDUMLA-HMT | MLP | 85.54 | 84.02 | 80.00 | 81.34 | 84.95 | 84.23 | 83.25 | 83.01 |
| | CLVQ | 90.24 | 89.00 | 88.04 | 87.35 | 89.77 | 88.25 | 87.65 | 86.20 |
| | CRBF | 86.52 | 85.35 | 80.56 | 79.54 | 85.36 | 81.26 | 83.01 | 82.14 |
| CASIA | MLP | 85.25 | 77.65 | 75.54 | 81.50 | 87.12 | 78.24 | 77.39 | 80.64 |
| | CLVQ | 89.35 | 88.25 | 88.26 | 88.36 | 88.98 | 88.03 | 88.54 | 88.02 |
| | CRBF | 82.64 | 80.25 | 79.65 | 77.24 | 81.24 | 80.54 | 80.65 | 79.54 |
| Our Collected Database | MLP | 90.24 | 90.20 | 88.32 | 88.74 | 91.31 | 90.32 | 90.54 | 88.55 |
| | CLVQ | 91.54 | 89.24 | 89.67 | 88.64 | 91.15 | 88.75 | 88.64 | 89.64 |
| | CRBF | 90.54 | 85.54 | 83.65 | 79.25 | 90.08 | 87.14 | 82.54 | 88.17 |

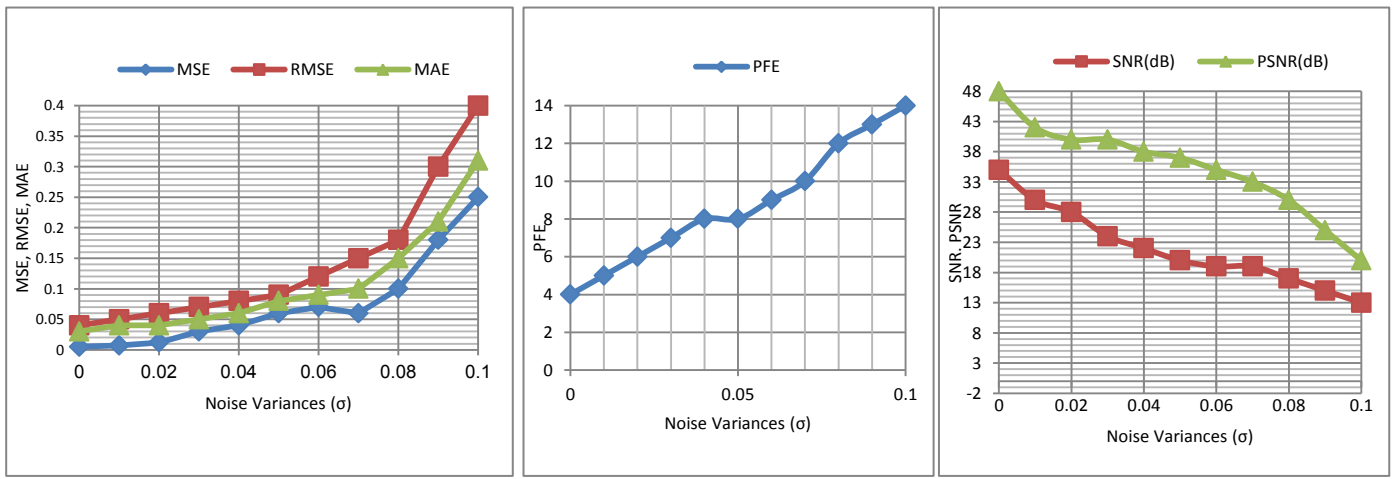

(a)　　　　　　　　　　　　(b)　　　　　　　　　　　　(c)

Fig 4: Performance metrics of the noise face images based on mean filter using σ = [0 0.1] for our collected database. (a) MSE, RMSE, and MAE. (b) PFE. (c) SNR, and PSNR.

TABLE 4: THE GAR (%) OF THE FACE IMAGES WITH ADDITIVE WHITE GAUSSIAN NOISE APPLIED ON OUR COLLECTED DATABASE

| SURF + | GAR (%) | | | |
|---|---|---|---|---|
| | MV | W. Sum | Prod | B. Count |
| MLP | **78.02** | 77.60 | 72.25 | 73.25 |
| CLVQ | **84.15** | 82.97 | 80.54 | 79.64 |
| CRBF | **80.36** | 82.64 | 77.58 | 80.00 |

## V. CONCLUSION AND FUTURE WORK

Although, face recognition systems are more efficient for human recognition including security fields, there are so many serious attempts to overcome such face recognition problems. These problems including illumination, poses, facial expression, aging, Botox, and noisy face images due to low quality camera. In this paper, we proposed a multi-view face recognition system based on SURF algorithm and neutral network approaches. The proposed system is evaluated using multi-layer perceptron (MLP), combined learning vector quantization (CLVQ), and combined radial basis function (CRBF) based on rank and decision level fusion. The decision is tested using majority voting (MV), weighted sum (W. Sum), product (Prod), and Borda count (B. Count). In this paper, we tested our system using invariant face images with different poses and angles using three databases SDUMLA-HMT, CASIA, and our collected datasets. By adding white Gaussian noise with zero means and variance (σ) we tested our collected database and a satisfactory results are achieved. In the future, we plan to use both thermal and video face images based on multi-sensor fusion to tackle the illumination problems. The occlusion, and patch problems also can be handled using convolutional deep neural networks techniques.

**Biography**

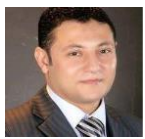
**Mahmoud Shams** is a Ph.D. candidate in the Computer Science Department, Mansoura University. He received a bachelor's degree in electronics and communication from the Faculty of Engineering, Mansoura University in 2004 and a master's degree in computer vision and pattern recognition from the Faculty of Computer and Information Sciences, Mansoura University in 2012. His research interests include deep learning approaches in multimodal biometric systems.

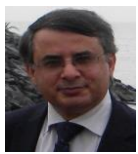
**Ahmad Tolba** is a professor of computer science at Mansoura University, Egypt. Professor Tolba holds a Ph.D. in computer vision from Wuppertal University in Germany and is the author of over 70 papers published in refereed international journals. He has served as director of the national project "ICT in Higher Education Development in Egyptian Universities." He has also served as a reviewer for various international journals, such as IEEE Transactions on Pattern Analysis and Machine Intelligence, Image Processing and Vision Computing, Digital Signal Processing, and Machine Vision and Applications.

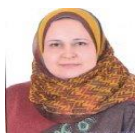
**Shahenda Sarhan** is a lecturer of computer science, Faculty of Computer and Information Sciences, Mansoura University. She received a Ph.D. in June 2012 and was awarded a prize for the best Ph.D. thesis in Mansoura University in 2014. She has published more than 20 papers in computer science and artificial intelligence in refereed international journals.